\documentclass[final]{cvpr}

\usepackage{times}
\usepackage{epsfig}
\usepackage{graphicx}
\usepackage{amsmath}
\usepackage{amssymb}

\usepackage[switch]{lineno} 
\usepackage{booktabs}
\usepackage{caption}
\usepackage{{makecell}}
\usepackage{amsmath}
\usepackage{xcolor}

\usepackage{varwidth}
\DeclareCaptionFormat{myformat}{%
  \begin{varwidth}{\linewidth}%
    \centering
    #1#2#3%
  \end{varwidth}%
}

\usepackage[pagebackref=true,breaklinks=true,colorlinks,bookmarks=false]{hyperref}

\begin{document}

\title{Discriminative Appearance Modeling with Multi-track Pooling for Real-time Multi-object Tracking}

\author{Chanho Kim$\,{}^{1}$ \qquad Li Fuxin$\,{}^{2}$ \qquad Mazen Alotaibi$\,{}^{2}$ \qquad James M. Rehg$\,{}^{1}$ \\[5pt] 
$^1$Georgia Institute of Technology \qquad $^2$Oregon State University 
}

\maketitle

\begin{figure*}
    \includegraphics[width=\textwidth]{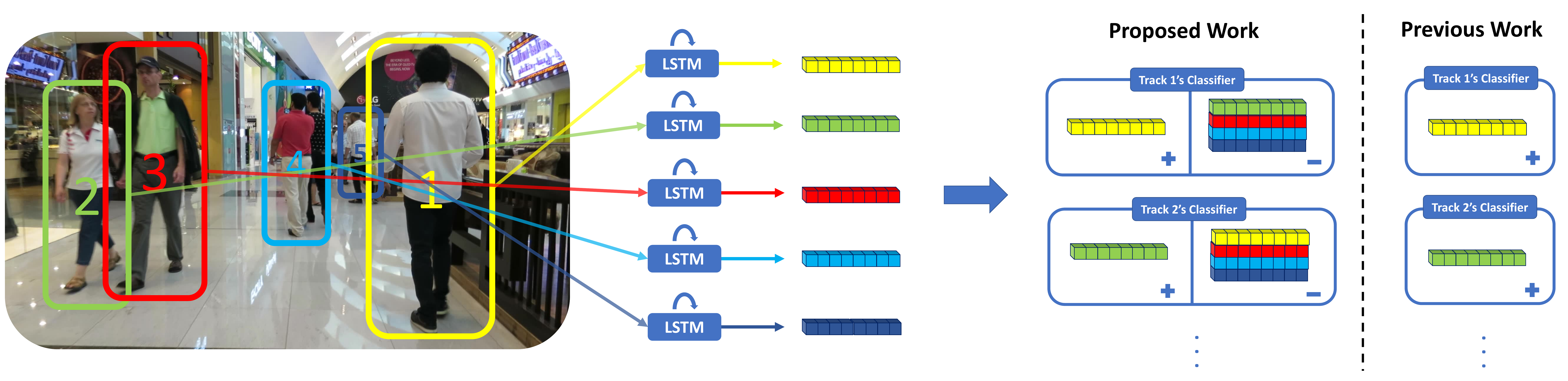}
    \caption{Existing recurrent neural network-based track classifiers used only matched detections for updating its appearance memory during tracking. This does not consider other objects in the scene (i.e. negative examples), which may have similar appearances. We propose to improve the predicted likelihood of such a classifier by augmenting its memory with appearance information about other tracks in the scene with multi-track pooling, leveraging the appearance information from the full set of tracks in the scene. The resulting classifier learns to adapt its prediction based on the information from other tracks in the scene.} \label{fig:method}
\end{figure*}

\begin{abstract}
In multi-object tracking, the tracker maintains in its memory the appearance and motion information for each object in the scene. This memory is utilized for finding matches between tracks and detections and is updated based on the matching result. Many approaches model each target in isolation and lack the ability to use all the targets in the scene to jointly update the memory. This can be problematic when there are similar looking objects in the scene. In this paper, we solve the problem of
simultaneously considering all tracks during memory updating, with only a small spatial overhead, via a novel multi-track pooling module. 
We additionally propose a training strategy adapted to multi-track pooling which generates hard tracking episodes online. 
We show that the combination of these innovations results in a strong discriminative appearance model, enabling the use of greedy data association to achieve online tracking performance. Our experiments demonstrate  real-time, state-of-the-art performance on public multi-object tracking (MOT) datasets. The code and trained models will be released at \url{https://github.com/chkim403/blstm-mtp}.
\end{abstract}

\section{Introduction}
In the typical tracking-by-detection setting of multi-object tracking, a trained target detector is assumed to exist, and the goal of the tracking algorithm is to solve the \textit{data association} problem: associating detections from different frames into tracks. 
The standard approach involves building an appearance model and a motion model for each target being tracked. The appearance model stores features from past appearances, which are compared against detections in each new frame, and the motion model predicts where the target will be located in the next frame. Because of the ambiguity in appearance, in the past, many high-performance trackers utilized sophisticated data association schemes that relied on evidence from multiple frames to determine the correct track. This has included frames from the future, sometimes with a significant look-ahead. These choices result in a tracker which is not capable of \textit{real-time online} performance, i.e. it  cannot generate a result \textit{without delay} after processing a new frame in the video. This is unfortunate, because real-time online tracking is critically important for numerous practical applications in AI. For example, in robotics applications such as autonomous driving, online processing is critical to support real-time decision-making.

Simple online tracking solutions based on matching appearance features between adjacent frames do not yield good performance. A better strategy 
is to use multiple frames to build up a memory~\cite{Sadeghian2017, fang2018recurrent, ckim2018} which has the ability to store multiple appearances and match them to new detections automatically. However, when matching the current detection and object tracks, most current approaches look at tracks one at a time without considering them jointly. This approach does not have sufficient discriminative power to address the ambiguities arising when several similar targets move adjacent to each other in the video, e.g. multiple men in black suits. In this situation, either more subtle features need to be utilized for target discrimination, or the likelihood of the matching needs to be decreased to accommodate this uncertainty. This in turn requires the matching approach to utilize the appearance information \emph{from other nearby tracks} in making a determination. The pairwise appearance models used in prior tracking methods lack the ability to make this comparison.

This paper introduces a novel approach to appearance modeling that takes \emph{all} of the tracked objects into account when matching detections. Suppose each track has a stored online memory about all of its past appearances, we propose a novel \emph{multi-track pooling module} which stores a max-pooled version of the memory from all other tracks. This extension allows the appearance model to take into account \textbf{online} negative examples coming from other objects within the same video (see Fig. \ref{fig:method}). We show that multi-track pooling greatly enhances the discriminative power of the appearance model for track scoring and improves overall performance.

We leverage the additional discriminative power of our matching approach to achieve online tracking performance by means of a simple, greedy data association method: Each track is matched with the detection that has the highest likelihood of belonging to the track. Such an association strategy is extremely efficient, resulting in fast performance. In this work we test the hypothesis that our discriminative appearance and motion modeling can enable this simple data association mechanism to achieve strong tracking performance.

Our online tracking framework additionally incorporates four components from recent tracking works. First, we utilize the Bilinear LSTM framework~\cite{ckim2018} as the basis for our matching approach, due to its effective memory construction. Second, we incorporate training strategies that handle long tracks using truncated backpropagation through time~\cite{ilya_phd_thesis}, in contrast to prior works~\cite{Sadeghian2017, ckim2018} in which the models were trained with short tracks. Third, we extend object tracks to frames without detections using a motion model, thereby compensating for missing detections. Fourth, we trained a bounding box coordinate corrector to correct bounding box coordinates, which is especially helpful for the extended tracks.

In summary, this paper makes the following contributions: 

\begin{enumerate}
  \item A novel multi-track pooling module which enables a track classifier to take online negative examples into account during testing and thus adjust its prediction adaptively depending on the objects in the scene.
  \item A training strategy that is adapted to train the proposed track pooling module by utilizing within-batch dependencies among tracks and that enables long sequence training with the original full tracks instead of short, random tracks used in \cite{ckim2018}.
  \item A real-time, online multi-object tracking algorithm that achieves state-of-the-art performance on standard tracking benchmarks.
\end{enumerate}

\section{Related Work}

Two groups of prior works have explored the incorporation of positive and negative samples during on-the-fly testing, and these are the closest related works to this paper. One line of work incorporates these samples by fine-tuning a pretrained CNN using
positive examples (target object) and negative examples (background in \cite{nam2016mdnet},  other objects in \cite{MaTang:ACCV:2018}) during testing. While these approaches share our interest in utilizing scene-specific information during tracking, the need to fine-tune during testing adds an additional source of complexity and is a barrier to efficient online performance. In contrast, our model automatically adjusts its prediction based on scene-specific information without the need for fine-tuning. The second line of work uses relational networks or graph convolutional networks (GCNs) to incorporate the appearance of other detections in the same frame~\cite{xu2019spatial} and in neighboring frames~\cite{mot_neural_solver_2020_CVPR}  when computing the appearance features of each detection. However, \cite{mot_neural_solver_2020_CVPR} operates in a batch setting where the entire video is available, whereas our approach is online. In addition, our multi-track pooling method is significantly simpler and faster than relational networks or graph convolutional networks which require  multiple iterations of message passing.

We utilize the Bilinear LSTM architecture of~\cite{ckim2018} in developing our matching approach. We extend beyond this work in multiple ways, the primary difference being the introduction of a novel multi-object pooling approach which utilizes appearance information across tracks to significantly improve data association performance. We demonstrate that this makes it feasible to use a much simpler and more cost-effective matching algorithm following track scoring, achieving real-time multi-object tracking. 

\cite{Sadeghian2017} presented an LSTM-based track proposal classifier that integrates motion, appearance, and interaction cues.
\cite{Zhu_2018_ECCV} presented a new attention network where attentions are computed both spatially and temporally to predict binary labels for track proposals.
\cite{fang2018recurrent} proposed recurrent autoregressive networks where they learn generative models for both motion and appearance, and these generative models are used in data association to determine how likely it is that a detection belongs to a track.
The primary difference between our work and these prior approaches is that our model makes a prediction for a detection-track pair by taking appearances of all tracks into account.

\cite{PellegriniESG09} and \cite{choi_eccv12} exploited interactions between tracks in solving the multi-object tracking problem.
\cite{PellegriniESG09} incorporated a social behavior model into their tracking algorithm. 
The social behavior model is based on the assumption that each person moves in such a way as to avoid collisions with other people.
\cite{choi_eccv12} incorporated high-level human activity cues into their tracking algorithm by exploiting the fact that human activities can influence how people move in the scene.
In contrast to these works, our work focuses on incorporating multiple appearances from all tracks into our model in order to make it more discriminative.

\section{Track Proposal Classifier} \label{tr_cls_section}

Tracking-by-detection approaches in multi-object tracking evaluate multiple track proposals \cite{choi2015near, CKIM2015, Tang2017, Sadeghian2017, ckim2018} when finding correspondences between tracks and detections.
Track proposals are typically constructed by extending existing tracks from the previous frame with new detections in the current frame. 
Let us denote $T_l(t)$ as the $l^{\text{th}}$ track at time $t$. Let $s(l)$ be the starting frame of the $l^{\text{th}}$ track and 
$d^l_t$ be the detection selected by the $l^{\text{th}}$ track at time $t$. We then write the $l^{\text{th}}$ track at time $t$ as $T_l(t) =\{d^l_{s(l)},\: d^l_{s(l)+1},\:...\, ,\: d^l_{t-1},\: d^l_{t} \}$. 
Recurrent  networks are trained to output the following conditional probability:
\begin{equation} \label{eq:cond_prob1}
f(d_t, T_l(t-1); \theta) = p(d_t \in T_l(t) | T_l(t-1))
\end{equation}
where $f(\cdot)$ and $\theta$ represent a neural network and its learnable parameters respectively. The inputs to the neural network are the detection-track pair $(d_t, T_l(t-1))$.

\subsection{Bilinear LSTM}

Vanilla LSTM equations are defined as follows \cite{Hochreiter97}:
\begin{eqnarray}
&\mathbf{f}_t = \sigma(\mathbf{W}_f [\mathbf{h}_{t-1};\mathbf{x}_t]),  &\mathbf{i}_t = \sigma(\mathbf{W}_i [\mathbf{h}_{t-1}; \mathbf{x}_t]), \nonumber\\
&\mathbf{g}_t = \sigma(\mathbf{W}_g [\mathbf{h}_{t-1}; \mathbf{x}_t]),  &\mathbf{o}_t = \tanh(\mathbf{W}_o [\mathbf{h}_{t-1}; \mathbf{x}_t]), \nonumber\\
&\mathbf{c}_t = \mathbf{f}_t \circ \mathbf{c}_{t-1} + \mathbf{i}_t \circ \mathbf{g}_t, &\mathbf{h}_t = \mathbf{o}_t \circ \tanh(\mathbf{c}_t)
\label{eq:lstm}
\end{eqnarray}
where $[;]$ denotes concatenation of 2 column vectors.
In the sequence labeling problem, $\mathbf{h}_t$ stores the information about the sequence and is fed into additional fully-connected or convolutional layers 
to generate the output.
The input $\mathbf{x}_t$ and the LSTM memory $\mathbf{h}_{t-1}$ are combined by additive interactions in the above equations. 

There are two issues with this formulation. First, the \textit{matching} operation is usually more easily represented by multiplicative relationships instead of additive, an intuitive example being the inner product as a correlation metric. Second, it is difficult to store and clearly distinguish multiple different appearances in the same LSTM memory vector, but a track exhibiting multiple different appearances in different frames is very common in multi-target tracking. Bilinear LSTM solves these issues by introducing a new memory representation based on the multiplicative interaction between the input and the LSTM memory:
\begin{eqnarray}
&\mathbf{h}_{t-1} &= [\mathbf{h}_{t-1,1}^\top,  \mathbf{h}_{t-1,2}^\top,  ...,  \mathbf{h}_{t-1,r}^\top]^\top \nonumber \\
&\mathbf{H}_{t-1}^{\text{reshaped}} &= [\mathbf{h}_{t-1,1}, \mathbf{h}_{t-1,2}, ...,  \mathbf{h}_{t-1,r}]^\top \nonumber \\
&\mathbf{m}_{t} &= g(\mathbf{H}_{t-1}^{\text{reshaped}}\mathbf{x}_{t}) 
\label{eq:blstm}
\end{eqnarray} 
where $g(\cdot)$ is a non-linear activation function. A long vector $\mathbf{h}_{t-1}$ from LSTM is reshaped into a matrix $\mathbf{H}_{t-1}^{\text{reshaped}}$ before it being multiplied to the input $\mathbf{x}_{t}$. Hence, multiple memory vectors can be matched with the feature from the detection with an inner product. The new memory $\mathbf{m}_{t}$ is then used as input to the final layers to generate the output instead of $\mathbf{h}_{t}$. Note that the way that the LSTM memory $\mathbf{h}_{t-1}$ interacts with the input $\mathbf{x}_{t}$ is changed in Bilinear LSTM, while standard LSTM memory updates shown in Eq. (\ref{eq:lstm}) are used since the memory vector $\mathbf{h}_{t-1}$ is retained as in Eq. (\ref{eq:blstm}).

Bilinear LSTM bears some resemblance to the popular transformers model \cite{NIPS2017_7181} in natural language processing in that both utilize a multiplicative relationship between the sequence and a new token, but they have some important differences. In a transformer model, 
the inner product is taken between the key of a new token with \textit{all} previous tokens in the sequence. This makes them very memory-inefficient  and unsuitable for online operations that span hundreds of frames in multi-object tracking. In Bilinear LSTM, the memory size is fixed, and the memory is updated automatically using the LSTM updating rule. This has the effect of automatically grouping similar appearances into the same row of the memory $\mathbf{H}_{t-1}^{\text{reshaped}}$ in Eq. (\ref{eq:blstm}), so that the memory size does not grow linearly with respect to the sequence length. Hence, we believe that Bilinear LSTM is a suitable choice for online tracking in which multiple appearances for each track need to be stored.

\subsection{Application to Multi-object Tracking}
When Bilinear LSTM is used in multi-object tracking, each track $T_l(t-1)$ will have its own LSTM memory $\mathbf{h}^l_{t-1}$ which is stored during the tracking process. All new detections at frame $t$ go through a CNN to generate their corresponding $\mathbf{x}_t$, which are then used to compare with $\mathbf{h}^l_{t-1}$ for all the tracks.  
When each detection has been scored with each track, the detections will be assigned to the existing tracks by either greedy assignment or multiple hypothesis tracking (MHT) assignment~\cite{CKIM2015}. Finally, the features from the assigned bounding box are used as $\mathbf{x}_t^l$ to update the track memory with Eq. (\ref{eq:lstm}). The updated memory $\mathbf{h}^l_{t-1}$ will be stored and then the same process will be repeated in the next frame. All tracks share the same LSTM network as their appearances will be dynamically updated in the memory, hence there is no re-training needed for any new object.

\subsection{Multi-Track Pooling Module}

\begin{figure}[t]
\centering
\includegraphics[width=1.0\columnwidth]{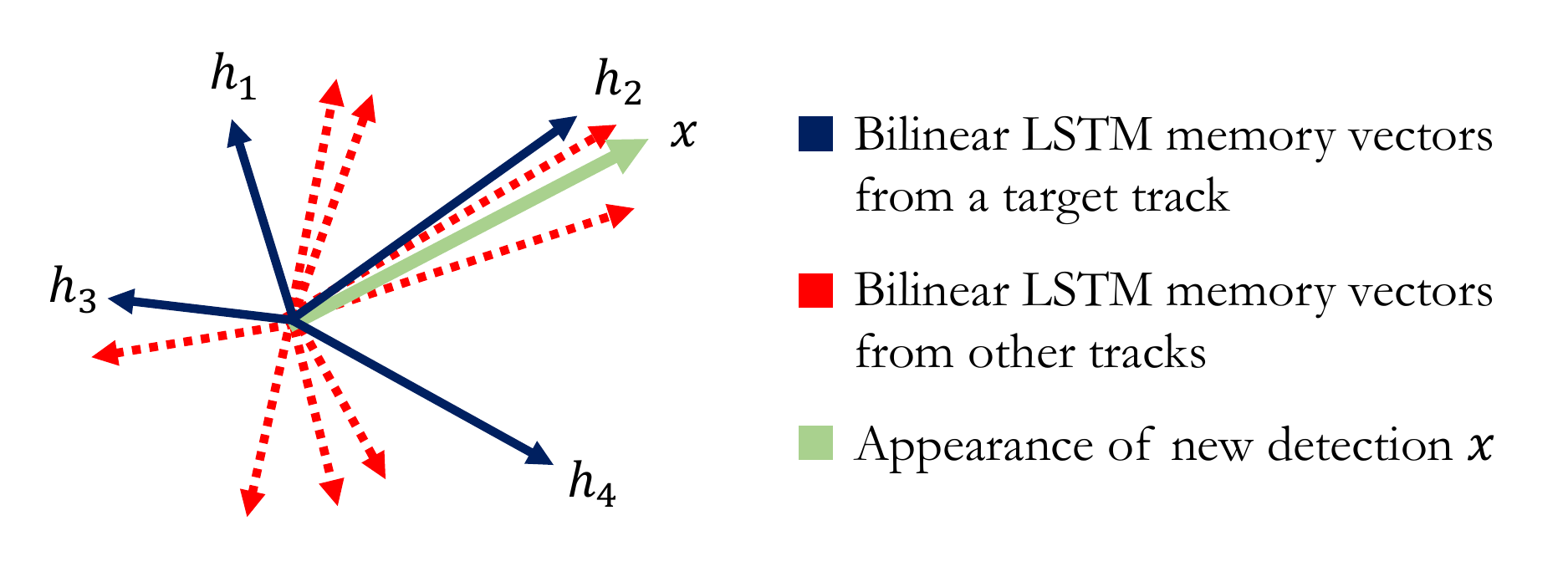}
\caption{Bilinear LSTM and the proposed improvements by multi-track pooling module. Bilinear LSTM stores multiple memory vectors for each track (in blue) so that a new detection $x_t$ can be matched with multiple templates via inner products. In this work, we improve on it by concatenating the memory with multiple stored memory vectors from other simultaneously tracked targets (in red) to serve as negative examples, so that the template matching process can take into account more subtle differences between the positive track and negative tracks (best viewed in color).}
\label{fig:blstm}
\end{figure}

A limitation of the previous work \cite{Sadeghian2017,ckim2018} is that only past appearances of the same track were considered, as only matched detections were inputted as $\mathbf{x}_t^l$ to update the LSTM memory at time $t$. 
However, during tracking, different targets can have similar appearances. For example, in pedestrian tracking, there could always be many people wearing similar white shirt and black pants, or black suits and black pants. These people need to be distinguished via more detailed features, such as shoes, objects in hand, etc. Thus, simple appearance matching may not be discriminative enough.

We propose to extend the track scoring process to consider \textbf{all} the tracked objects, other than only the current object of concern. Instead of simply growing the memory to store more templates, which can hit a memory bottleneck, we jointly consider all the objects that have been tracked. Tracked objects are usually the ones most easily confused with the target (if a detection does not come from a track, then its appearance is likely significantly different from any tracked objects (e.g. pedestrians)). Hence, taking the appearance information about these objects into consideration could greatly improve the discriminative power of the tracker for each target (see Fig. \ref{fig:blstm}).

In order to consider other tracked objects as well, we propose to modify Eq. (\ref{eq:cond_prob1}) to:
\begin{equation} \label{eq:cond_prob2}
f(d_t, T_{1:M}(t-1); \theta) = p(d_t \in T_l(t) | T_{1:M}(t-1))
\end{equation}
where $T_{1:M}(t-1)=\{T_1(t-1), T_2(t-1), ..., T_M(t-1)\}$ represents the existing tracks in the previous frame. In this paper, we denote $M$ as the number of tracks in the previous frame $t-1$ and denote $N$ as the number of detections in the current frame  $t$. \cite{Sadeghian2017} also trained recurrent neural networks to output this kind of conditional probability by using track interaction cues, but their interaction cues were the most recent locations of other tracks, not appearances of other tracks.

In Bilinear LSTM, 
each track in the previous frame is associated with a unique LSTM memory $\mathbf{h}^l_{t-1}$. When new detections $\mathbf{x}_t$ arrive in the current frame, we compute $\mathbf{m}_{t}$ in Eq. (\ref{eq:blstm}) between each of the $M$ existing tracks and each of the $N$ new detections. We denote $\mathbf{m}_t^+$ as the matching between $\mathbf{h}^l_{t-1}$ computed using the target track (i.e. the current object of concern) and $\mathbf{x}_t$ (as computed by Eq. (\ref{eq:blstm})). For the other $M-1$ tracks in the scene, we denote $\mathbf{M}^{-}_t$ as a matrix in which each row represents $\mathbf{m}_{t}$ computed by a non-target object track (i.e. other objects in the scene) and the new detection ($\mathbf{x}_t$):

\begin{equation} 
\mathbf{M}^{-}_{t} = [\mathbf{m}^{-}_{t,1},  \mathbf{m}^{-}_{t,2}, ... , \mathbf{m}^{-}_{t,M-1}]^\top 
\end{equation}
where $\mathbf{m}^{-}_{t,i}$ represents the matching between the $i$-th non-target object track and the new detection.

Here the main difficulty is that we have one positive track and an indefinite number of negative tracks. We propose to compress $M - 1$ $\mathbf{m}_{t}$s using max pooling in order to obtain a fixed size memory representation $\mathbf{m}^{-}_{t}$ regardless of the number of the tracks in the previous frame:

\begin{equation} \label{eq:neg_mem}
\mathbf{m}^{-}_{t}(j) = \max_{i} \mathbf{M}^{-}_{t}(i,j)
\end{equation}
where $\mathbf{m}^{-}_{t}(j)$ represents the $j$-th element in $\mathbf{m}^{-}_{t}$, and $\mathbf{M}^{-}_{t}(i,j)$ represents an element located in the $i$-th row and $j$-th column of $\mathbf{M}^{-}_{t}$. Thus, the $j$-th element in $\mathbf{m}^{-}_{t}$ is computed by taking the maximum of the $j$-th column of $\mathbf{M}^{-}_{t}$.

Since the Bilinear LSTM memory stores correlation responses between multiple templates  and the input, applying max pooling to $\mathbf{M}^{-}_{t}$ allows us to detect high correlation responses generated by non-target object tracks for the detection which is currently considered. Thus, values in $\mathbf{m}^{-}_{t}$ can show whether any of the non-target object tracks has a similar appearance to that of the current detection. 
We obtain the final memory representation for a track classifier by concatenating $\mathbf{m}^{+}_t$ and $\mathbf{m}^{-}_t$ as:

\begin{equation} 
\mathbf{m}^{\text{all}}_{t} = [\left( \mathbf{m}^{+}_{t} \right); \left( \mathbf{m}^{-}_{t} \right)]
\end{equation}
The network runs a fully-connected layer after  $\mathbf{m}^{\text{all}}_{t}$ followed by softmax,  that outputs the binary decision that determines whether the new detection belongs to the target  track ($\mathbf{m}^{+}_{t}$).

\begin{table}
\centering
\scalebox{0.43}{
    \begin{tabular}{|lr||lr||lr||lr|}
    \hline
    \multicolumn{8}{ |c| }{Soft-max \qquad \qquad \quad \quad \ \, \, 2} \\
    \hline
    \multicolumn{8}{ |c| }{FC \qquad \qquad \qquad \qquad \ \ \ \, \, 2 } \\
    \hline
    \multicolumn{8}{ |c| }{Concatenation ($\mathbf{m}^{\text{all}}_{t}$) \qquad 16 } \\
    \hline
    \multicolumn{4}{ |c|| }{Matrix-vector Multiplication-relu ($\mathbf{m}^{+}_t$) \quad \quad \quad  \quad  \quad \quad \quad 8} & \multicolumn{4}{ c| }{Max-pooling ($\mathbf{m}^{-}_t$) \quad \quad \quad  \quad  \quad \quad \quad \quad \quad \quad \quad \quad  \quad \ \quad  \quad \quad \quad \quad \, \ \ \ \ \ \ $8$} \\
    \hline
    \multicolumn{2}{ |c|| }{} & \multicolumn{2}{ c|| }{} & \multicolumn{4}{ c| }{Matrix-vector Multiplication-relu ($\mathbf{M}^{-}_t$) \quad \quad \quad  \quad  \quad \quad \quad $(M-1) \times 8$} \\
    \hline
    Reshape & $8 \times 256$ &Reshape &  $256 \times 1$ & Reshape & $(M-1) \times 8 \times 256$ &Reshape &  $256 \times 1$\\ 
    \hline
    LSTM & 2048  & & & LSTM & $(M-1) \times 2048$  & &\\
    \hline
    FC-relu & 256 & FC-relu & 256 & FC-relu & $(M-1) \times 256$ & FC-relu & 256  \\
    \hline
    ResNet-50 & 2048 & ResNet50 & 2048 & ResNet-50 & $(M-1) \times 2048$ & ResNet50 & 2048 \\
    \hline
    $\mathbf{x}^{+}_{t-1}$ & $128 \times 64 \times 3$ &  $\mathbf{x}_{t}$ & $128 \times 64 \times 3$ & $\mathbf{x}^{-}_{t-1, i}$ & $(M-1) \times 128 \times 64 \times 3$ &  $\mathbf{x}_{t}$ & $128 \times 64 \times 3$ \\
    \hline
    \end{tabular}
}
\caption{Proposed Network Architecture for Track Proposal Classifier. The right two columns represent the multi-track pooling module where ($M -1$) other tracks are processed. %
}
\label{table:architecture}
\end{table}

Table \ref{table:architecture} shows the proposed network architecture. In practice we compute the Bilinear LSTM memories for all the tracks and then construct $M_t^-$ by simply stacking the precomputed memory vectors into a matrix, which can be done efficiently. We adopt ResNet 50 \cite{He2015} as our convolutional neural network and use Bilinear LSTM \cite{ckim2018} as our recurrent network. In addition to the proposed appearance model, we adopt the motion model presented in \cite{ckim2018} as the learned motion model. It is a binary classifier based on a traditional LSTM that receives the bounding box coordinates as input. We combined the appearance model and motion model to form a joint model that utilizes both appearance and motion cues. We used this joint model to generate the results in Table \ref{table:mot17_tracktor}, \ref{table:mot17}, and \ref{table:mot16}. More details about the joint model's network architecture can be found in the supplementary. 

\section{Training}
\label{sec:training}

In this section, we describe the training strategy for the proposed neural network architecture by sampling mini-batches so that tracks within the mini-batch are correlated, in order for the multi-track pooling module to train effectively. 
We explain the method that generates our training data consisting of both \textit{actual} multi-object tracking episodes and \textit{random} tracking episodes from public multi-object tracking datasets \cite{MOT15, MOT16} and then present our choice of loss function.

\subsection{Actual Tracking Episodes as Training Data} \label{sec:actual_track_data}

We generate actual tracking episodes by processing ground truth track labels sequentially and generating track proposals every frame in an online manner. Specifically, when we have $M$ tracks in the previous frame and $N$ detections in the current frame,
we generate $MN$ track-detection pairs in the current frame by considering all possible pairs between the existing tracks and the new detections.
Each proposal is associated with a binary label where positive label indicates a match and negative represents  that the track and detection belong to different objects. 
We use these $MN$ track proposals as a mini-batch for each training iteration and repeat this process until we process all the frames in video. Then we move on to the next training video. 
This process is repeated over all training videos during training.

\textbf{Truncated backpropagation through time.} As we process training videos sequentially, tracks become too long to fit into GPU memory. 
One way to avoid this is to split tracks into shorter ones, which is similar to the training data used in \cite{Sadeghian2017, ckim2018} for recurrent models. 
However, this is sub-optimal because a recurrent model would only be able to learn temporal structures with an extent limited by the maximum sequence length that the model sees during training.

In order to address this issue, we adopt truncated backpropagation through time (BPTT) \cite{ilya_phd_thesis}. 
In truncated BPTT, the maximum number of time steps where the loss is backpropagated is limited by a fixed-size time window. 
However, entire sequences are processed by a recurrent model through multiple training iterations by a moving time window. Since long sequences are processed through multiple training iterations, the recurrent memories that store the sequence information need to be shared across the training iterations.
This allows the model to access the temporal context beyond the temporal extent which is determined by the size of the time window.
By training a recurrent model with the original long tracks, 
the model has a better chance to learn how to exploit long-term information when solving the track proposal classification problem.

\subsection{Random Tracking Episodes as Training Data} \label{sec:random_track_data}

We also use short random track proposals, which is similar to the training data used in \cite{Sadeghian2017, ckim2018}, as our additional training data. We generate short random track proposals as follows. At each training iteration, we first pick a video where we generate tracks and randomly select the start frame and end frame for obtaining track proposals. We take all the tracks that exist in the end frame. 
Denote $N_{\text{max}}$ as the maximum number of tracks that we use to construct random track proposals. 
If the number of the tracks in the end frame is greater than $N_{\text{max}}$, we randomly pick $N_{\text{max}}$ tracks among them. 
In order for each mini-batch to have tracks of different lengths, we randomly clip the selected tracks such that each track in the batch start in different frames.

We take $N$ detections from the selected tracks in the end frame and take $M$ tracks from the previous frame to form track proposals ($M < N$ when new tracks are born in the selected end frame. Otherwise, $M=N$). We generate $MN$ track proposals by considering all possible matchings between these two.
Thus, $M$ track proposals are associated with a positive label (same object),
and $MN - M$ proposals are associated with a negative label.

\subsection{Loss Function}

For each mini-batch, there are $MN$ track proposals, each of which incurs a loss value. 
Thus, our cross-entropy loss is written as:
\begin{equation} \label{eq:loss}
L(t) =  \frac{1}{MN}\sum_{i=1}^{M} \sum_{j=1}^{N}  \alpha_{ij}(t) L_{ij}(t) \\
\end{equation}
where $t$ represents a frame number in video, and $\alpha_{ij}$ is a weighting factor. The cross-entropy loss term for each training example $L_{ij}(t)$ is defined as:
\begin{equation}
\begin{split}
L_{ij}(t) = \qquad \qquad \qquad \qquad \qquad \qquad \qquad \qquad \qquad \qquad& \\
\begin{cases}
- &\log p(d_{jt} \in T_i(t) | T_{1:M}(t-1)), \text{if} y_{ij}(t)=1 \\
- &\log (1-p(d_{jt} \in T_i(t) | T_{1:M}(t-1))),  \text{otherwise}
\end{cases}
\end{split}
\end{equation}
where $d_{jt}$ represents a $j$th detection in $t$, and $y_{ij}(t)$ is a ground truth binary label representing whether $d_{jt}$ belongs to a track $T_i(t)$ or not. For the weighting factor $\alpha_{ij}(t)$, we adopted the weighting factor of Focal loss \cite{Lin2017b} 
to address the class imbalance of our training data (i.e. there are much more negative examples than positive examples). The weighting factor is then written as:
\begin{equation}
\begin{split}  \label{eq:weight_factor}
    \alpha_{ij}(t) = \qquad \qquad \qquad \qquad \qquad \qquad \qquad \qquad \qquad \qquad&  \\
    \begin{cases}
    \beta_+ (1 - p(d_{jt} \in T_i(t) | T_{1:M}(t-1)))^2, &\text{if } y_{ij}(t)=1 \\
    \beta_- (p(d_{jt} \in T_i(t) | T_{1:M}(t-1)))^2 , &\text{otherwise}
    \end{cases}
\end{split}
\end{equation} 
where $\beta_+$ and $\beta_-$ are class-specific constant weights which are found empirically as suggested in \cite{Lin2017b}. For positive labels, we used $\beta_+=4$, and, for negative labels (i.e. $d_{jt} \notin T_i(t)$), we used $\beta_-=1$.

\section{Tracking Algorithm}

In this work, we chose to use the greedy association because it is the fastest association algorithm, and our training setting simulates such a greedy association algorithm (i.e. the loss value was calculated for each track independently and then averaged across all tracks in a mini-batch). The greedy data association algorithm runs in $O(MN)$ time.
A more complicated scheme would require additional computational complexity (e.g. the Hungarian algorithm runs in $O(MN^2)$)  and would require more careful parameter tuning for the tracker. 

\subsection{Greedy Data Association} \label{sec:greedy_tracker}

Our greedy data-association works as follows. It initializes new tracks with the detections in the first frame. The tracker generates and stores the LSTM memory $\mathbf{h}_{t-1}$ for each of the tracks. When new detections $\mathbf{x}_t$ arrive in the next frame, we compute the association likelihood for every possible track-detection pair using the track proposal classifier. We set the threshold for the association likelihood to 0.5. The data association problem is then solved in a greedy manner starting from the highest matching likelihood. The tracks are updated with the newly assigned detections, and new tracks are born from the detections which are not associated with any of the existing tracks. The tracker updates the LSTM memory from $\mathbf{h}_{t-1}$ to  $\mathbf{h}_{t}$ for every track according to the data association result. This process is repeated until all the video frames are processed.

We adopt a simple track termination strategy which terminates tracks if they have more missing detections than actual detections or the number of consecutive missing detections becomes larger than a threshold $N_{\text{miss}}$. The terminated tracks will not be used in the data association process anymore and will not be used for computing non-target object memories either.

\subsection{Track Extension and Bounding Box Correction} \label{sec:track_extension}

In online tracking, we attempt to extend a track by generating additional detections. When there is no detection in a predicted target location in the current frame, we generate a new detection bounding box using the location predicted by a Kalman filter and then use our track proposal classifier to decide whether or not the newly generated detection belongs to the current target. Since we rely on motion cues when generating these additional detections, the bounding box coordinates of the newly generated detections might not be accurate. Thus, we train a bounding box coordinate corrector which predicts the correct bounding box coordinates based on the CNN features, which is similar to the bounding box regression module presented in \cite{girshick2014rcnn}. More details are included in the supplementary material. 

\section{Experiments}

We tested the proposed method on the MOT 16 and MOT 17 (Multiple Object Tracking) Challenge Benchmark \cite{MOT16}. We train on the MOT 17 training sequences, as well as  additional public training sequences including 7 MOT 15 training sequences \cite{MOT15} which were not included in MOT 17 training/testing sequences to our training data. In contrast to several recent work \cite{Tang2017, Sadeghian2017}, we did not utilize the Person Re-identification datasets \cite{li2014deepreid, zheng2015scalable} to pretrain our CNN.

For our ablation study, we used two validation sets. The first one is the MOT 19 Challenge training sequences \cite{MOT19_CVPR} which have 2,390 annotated object tracks. In contrast to the MOT 17 sequences, this new challenge dataset provides heavily crowded scenes captured in new environments, which makes it good for validating the proposed appearance model. Note that this dataset was only used for our ablation study and thus was not used to train our model in Table \ref{table:mot17_tracktor}, \ref{table:mot17}, and \ref{table:mot16}. The second validation set is the MOT 17 Challenge training sequences which have 512 annotated object tracks. When the second validation set is used, we used the MOT 15 training sequences as our training data. See the supplementary document for the video sequence names used in the training, validation, and test sets. 

We used the standard MOT metrics such as IDF1 \cite{Ergys2016}, IDS, MOTA \cite{Bernardin2008}, Mostly Tracked (MT), Mostly Lost (ML), and Fragmentation (Frag)~\cite{MOT16} for performance comparison in our experiments.

\subsection{Ablation Study}

In the ablation study, we show the effectiveness of the proposed multi-track pooling module by comparing its performance to the original Bilinear LSTM. Firstly, we tested their performance without using any motion cues during tracking. Secondly, we used motion cues during tracking by adopting a simple motion gating strategy that allows detections that are close to the current track to be considered as a possible matching. 
\begin{table} [!t]
	\centering
	\scalebox{0.83}{
	\begin{tabular}{ccccccccc}
	\toprule
	Method	&MOTA &IDF1 &IDS &MT &ML &Frag \\
	\midrule	
	B-LSTM  & 44.8  & 31.3  & 15,367 & 12.6 & 27.0 & 38,182\\	
	Ours  & 44.9 & \textbf{35.0}  & \textbf{11,940}  & 12.7 & 27.5 & \textbf{37,017}  \\
	\bottomrule
	\end{tabular}
	}

\caption{Performance comparison on MOT 19 train sequences (val1) when motion gating is not used.} \label{table:blstm_no_mot}
\end{table}

\begin{table} [!t]
	\centering
	\scalebox{0.86}{
	\begin{tabular}{ccccccccc}
	\toprule
	Method	&MOTA &IDF1 &IDS &MT &ML &Frag \\
	\midrule	
	B-LSTM  & 49.1  & 52.5  & 1,112 & 21.1 & 31.9 & 1,066\\	
	Ours  & 49.4 & \textbf{53.9}  & \textbf{809}  & 21.1 & 31.5 & 1,070  \\
	\bottomrule
	\end{tabular}
	}
\caption{Performance comparison on MOT 17 train sequences (val2) when motion gating is not used.} \label{table:blstm_no_mot_mot17}
\end{table}

\begin{table} [!t]
	\centering
	\scalebox{0.82}{
	\begin{tabular}{ccccccccc}
	\toprule
	Method	&MOTA &IDF1 &IDS &MT &ML &Frag \\
	\midrule	
	B-LSTM  & 45.1  & 39.6  & 9,137 & 12.6 & 27.6 & 36,379 \\	
	Ours  & 45.0 & \textbf{40.5}  & \textbf{7,873}  & 12.6 & 28.1 & \textbf{35,169}  \\
	\bottomrule
	\end{tabular}
	}
\caption{Performance comparison on MOT 19 train sequences (val1) when the simple motion gating strategy is used.} \label{table:blstm_simple_mot}
\end{table}

\begin{table} [!t]
	\centering
	\scalebox{0.86}{
	\begin{tabular}{ccccccccc}
	\toprule
	Method	&MOTA &IDF1 &IDS &MT &ML &Frag \\
	\midrule	
	B-LSTM  & 49.3  & 56.7  & 847 & 21.1 & 32.1 & 1,038 \\	
	Ours  & 49.6 & 56.8  & \textbf{616}  & 21.1 & 32.1 & 1,040  \\
	\bottomrule
	\end{tabular}
	}
\caption{Performance comparison on MOT 17 train sequences (val2) when the simple motion gating strategy is used.} \label{table:blstm_simple_mot_mot17}
\end{table}

\begin{table*}[!t]
\centering
\scalebox{0.84}{
\begin{tabular}{ccccccccccc}
\midrule
Method	& {Type} &IDF1 & MOTA &IDS &MT &ML &Frag &FP &FN & Hz\\
\midrule

GSM-Tracktor \cite{ijcai2020-74} & {online}   & \textcolor{blue}{57.8} & \textbf{56.4} & \textcolor{blue}{1,485} & \textbf{22.2}  & \textbf{34.5} & \textbf{2,763} & 14,379  & \textbf{230,174} & \textcolor{blue}{8.7} \\
Tracktor++v2 \cite{Philipp19} & {online}  & 55.1 & \textcolor{blue}{56.3} & 1,987 & \textcolor{blue}{21.1}  & \textcolor{blue}{35.3} & \textcolor{blue}{3,763} & \textcolor{blue}{8,866}  & \textcolor{blue}{235,449} & 1.5 \\
TrctrD17 \cite{xu2020train} & {online}  & 53.8 & 53.7 & 1,947 & 19.4  & 36.6 & 4,792 & 11,731  & 247,447 & 4.9 \\
Tracktor++ \cite{Philipp19} & {online}  & 52.3 & 53.5 & 2,072 & 19.5  & 36.6 & 4,611 & 12,201  & 248,047 & 1.5 \\
\textbf{Ours} & {online}  & \textbf{60.4} & 55.9 & \textbf{1,188} & 20.5 & 36.7 & 4,187  & \textbf{8,653}  & 238,853 & \textbf{24.8} \\
\midrule
\end{tabular}
}
\centering
\captionsetup{format=myformat}
\caption{MOT 17 Challenge (with trackers that utilized public detections + Tracktor \cite{Philipp19}). \newline Note that we used \textbf{bold} for the best number and \textcolor{blue}{blue} color for the second-best number.} \label{table:mot17_tracktor}
\end{table*}

\begin{table*}[!t]
\centering
\scalebox{0.80}{
\begin{tabular}{ccccccccccc}
\midrule
Method	& {Type} &IDF1 & MOTA &IDS &MT &ML &Frag &FP &FN & Hz\\
\midrule

STRN-MOT17 \cite{xu2019spatial} & {online}  & \textbf{56.0} & 50.9 & 2,397 & 18.9  & \textcolor{blue}{33.8} & 9,363 & 25,295  & \textcolor{blue}{249,365} & 13.8 \\
DMAN \cite{Zhu2018} & {online}  & \textcolor{blue}{55.7} & 48.2 & \textcolor{blue}{2,194} & \textcolor{blue}{19.3}  & 38.3 & 5,378 & 26,218  & 263,608 & 0.3\\
MOTDT17 \cite{long2018tracking} & {online}  & 52.7 & 50.9 & 2,474 & 17.5  & 35.7 & 5,317 & 24,069 & 250,768 & \textcolor{blue}{18.3}\\
AM-ADM17 \cite{SLee2018} & {online}  & 52.1 & 48.1 & 2,214 & 13.4  & 39.7 & \textcolor{blue}{5,027} & 25,061  & 265,495 & 5.7\\
HAM-SADF17 \cite{Yoon2018} & {online}   & 51.1 & 48.3 & \textbf{1,871} & 17.1  & 41.7 & \textbf{3,020} & \textcolor{blue}{20,967}  & 269,038  & 5.0\\
PHD-GSDL17 \cite{Fu2018ParticlePF} & {online} & 49.6 & 48.0 & 3,998 & 17.1  & 35.6 & 8,886 & 23,199 & 265,954 & 6.7\\
FAMNet \cite{Chu2019} & {online}   & 48.7 & \textbf{52.0} & 3,072 & 19.1  & \textbf{33.4} & 5,318 & \textbf{14,138}  & 253,616 & 0.0\\
\textbf{Ours} & {online}  & 54.9 & \textcolor{blue}{51.5} & 2,563 & \textbf{20.5} & 35.5 & 7,745  & 29,623  & \textbf{241,618} & \textbf{20.1} \\
\midrule
MHT-bLSTM \cite{ckim2018} & {near-online} & \textcolor{blue}{51.9} & 47.5 & \textcolor{blue}{2,069} & 18.2  & 41.7 & 3,124 & 25,981  & 268,042 & \textcolor{blue}{1.9} \\
EDMT17 \cite{Chen2017EnhancingDM} & {near-online}  & 51.3 & 50.0 & 2,264 & \textcolor{blue}{21.6}  & \textcolor{blue}{36.3} & 3,260 & 32,279  & \textcolor{blue}{247,297}  & 0.6\\
MHT-DAM \cite{CKIM2015} & {near-online}   & 47.2 & \textcolor{blue}{50.7} & 2,314 & 20.8  & 36.9 & \textcolor{blue}{2,865} & \textbf{22,875} & 252,889 & 0.9  \\
\textbf{Ours} & {near-online} & \textbf{55.8} & \textbf{53.6} & \textbf{1,845} & \textbf{23.4} & \textbf{34.5} & \textbf{2,299}  & \textcolor{blue}{23,669}  & \textbf{236,226} & \textbf{22.7} \\
\midrule
\end{tabular}
}
\caption{MOT 17 Challenge (Published online and near-online methods using public detections). } \label{table:mot17}
\end{table*}

\begin{table*}[!t]
\centering
\scalebox{0.83}{
\begin{tabular}{ccccccccccc}
\midrule
Method	& {Type} &IDF1 & MOTA &IDS &MT &ML &Frag &FP &FN & Hz\\
\midrule

STRN-MOT16 \cite{xu2019spatial} & {online}  & \textcolor{blue}{53.9} & \textbf{48.5} & 747 & \textcolor{blue}{17.0}  & \textbf{34.9} & 2,919 & 9,038  & \textcolor{blue}{84,178} & 13.5 \\
DMAN \cite{Zhu2018} & {online}  & \textbf{54.8} & 46.1 & \textcolor{blue}{532} & \textbf{17.4}  & 42.7 & 1,616 & 7,909 & 89,874 & 0.3 \\
MOTDT \cite{long2018tracking} & {online}  & 50.9 & 47.6 & 792 & 15.2  & 38.3 & 1,858 & 9,253 & 85,431 & \textcolor{blue}{20.6}\\
STAM16 \cite{Chu2017OnlineMT} & {online}  & 50.0 & 46.0 & \textbf{473} & 14.6  & 43.6 & \textcolor{blue}{1,422} & 6,895 & 91,117 & 0.2\\
RAR16pub \cite{fang2018recurrent} & {online}  & 48.8 & 45.9 & 648 & 13.2  & 41.9 & 1,992 & 6,871 & 91,173 & 0.9 \\
KCF16 \cite{Peng2019} & {online}  & 47.2 & \textcolor{blue}{48.8} & 648 & 15.8  & \textcolor{blue}{38.1} & \textbf{1,116} & \textcolor{blue}{5,875} & 86,567 & 0.1 \\
AMIR \cite{Sadeghian2017} & {online}  & 46.3 & 47.2 & 774 & 14.0  & 41.6 & 1,675 & \textbf{2,681} & 92,856 & 1.0 \\
\textbf{Ours} & {online}  & 53.5 & 48.3 & 733 & \textcolor{blue}{17.0} & 38.7 & 2,349  & 9,799  & \textbf{83,712} & \textbf{21.0} \\
\midrule
NOMT \cite{choi2015near} & {near-online}  & \textbf{53.3} & \textcolor{blue}{46.4} & \textbf{359} & \textcolor{blue}{18.3}  & 41.4 & \textbf{504} & 9,753 & \textcolor{blue}{87,565} & \textcolor{blue}{2.6}\\
EDMT \cite{Chen2017EnhancingDM} & {near-online}  & 47.9 & 45.3 & 639 & 17.0  & \textcolor{blue}{39.9} & 946 & 11,122  & 87,890  & 1.8\\
MHT-bLSTM \cite{ckim2018} & {near-online} & 47.8 & 42.1 & 753 & 14.9  & 44.4 & 1,156 & 11,637  & 93,172  & 1.8\\
MHT-DAM \cite{CKIM2015} & {near-online}   & 46.1 & 45.8 & 590 & 16.2  & 43.2 & 781 & \textbf{6,412} & 91,758  & 0.8 \\
\textbf{Ours} & {near-online} & \textcolor{blue}{52.5} & \textbf{49.9} & \textcolor{blue}{579} & \textbf{19.7} & \textbf{38.6} & \textcolor{blue}{674}  & \textcolor{blue}{7,111}  & \textbf{83,676} & \textbf{23.8} \\
\midrule
\end{tabular}
}
\caption{MOT 16 Challenge (Published online and near-online methods using public detections).} \label{table:mot16}
\end{table*}

\textbf{Data Association.} We ran the greedy data association algorithm described in the previous section with the following hyperparameter setting: 0.5 as the association threshold and $N_{\text{miss}}=60$.
In the ablation study, we did not interpolate missing detections in the final tracks using our track extension module in order to make the MOTA scores close across different methods. This allowed us to compare other tracking metrics in a fairer setting.

\textbf{Comparison with Bilinear LSTM.}
We examined the effect of the proposed multi-track pooling module on the tracking performance. For this ablation study to be fair, both Bilinear LSTM and our method were trained on the same training set (see the supplementary document) with the same training setup. Table \ref{table:blstm_no_mot} and \ref{table:blstm_no_mot_mot17} show the tracking results when no motion cues were used during tracking. In this case, the appearance model needed to do the heavy lifting. Table \ref{table:blstm_simple_mot} and \ref{table:blstm_simple_mot_mot17} show the tracking results when a simple motion gating strategy was applied. 
Bilinear LSTM with the multi-track pooling module consistently outperformed the original Bilinear LSTM on IDF1, IDS and Fragmentations, showing its effectiveness in multi-object tracking on both of our validation sets.

\subsection{MOT Challenges}

We evaluated both the online and near-online versions of our tracker for the MOT 17/16 Benchmarks. In the online version, we utilized the track extension module described in the previous section to recover missing detections (except for the tracker in Table \ref{table:mot17_tracktor} in which we turned off both the extension module and the bounding box corrector). In the near-online version, we performed local track smoothing instead for recovering missing detections. For the second case, we denote the method as near-online in Table \ref{table:mot17} and \ref{table:mot16} since local track smoothing requires lookahead frames. 

Recent approaches \cite{xu2020train, ijcai2020-74} utilized Tracktor \cite{Philipp19} to first refine public detections, which resulted in higher scores due to more accurate detections. In order to compare with these recent approaches, we also used public detections processed by Tracktor as input to our tracker and presented the comparison in Table \ref{table:mot17_tracktor} separately. It can be seen that our approach significantly improves the IDF1 score and identity switches over other online tracktor-based approaches, besides being at least 3 times faster than the nearest competitor. Note that this does not include the processing time spent by Tracktor.

In Table \ref{table:mot17} and \ref{table:mot16}, we did not utilize Tracktor and compared our results with other online and near-online trackers which did not utilize Tracktor. Again our greedy tracker is the fastest among the top performing trackers and our performance is comparable with the best trackers.  Considering its simplicity and speed, we believe our method demonstrates strong state-of-the-art performance on the MOT Challenge.

In near-online trackers, we significantly improve over our baseline MHT-bLSTM on both IDF1 (by $7.5\%$) and MOTA (by $12.8\%$), obtaining the best performance in near-online tracking. We also have the smallest amount of identity switches and fragmentations and the most objects that are mostly tracked in MOT 17. 
Note that these are obtained with a greedy data association algorithm and only local smoothing is added on top of the online tracker performance, hence the speed is even faster than the online version since local smoothing removed more false positives.  

\section{Conclusion}

In this paper, we introduce a novel multi-track pooling module that enables joint updating of appearance models using all tracks, thereby improving matching reliability when targets are similar in appearance. We propose a novel training strategy for track pooling that utilizes within-batch dependencies among tracks and supports training over long sequences. The resulting tracker is based on a Bilinear LSTM architecture and performs greedy data association. With this simple approach, it achieves real-time tracking performance with an accuracy equivalent to state-of-the-art trackers that are significantly slower.

{\small
\bibliographystyle{ieee_fullname}
\bibliography{egbib}
}

\clearpage

\section*{Supplementary Material}
\appendix

\section{Additional Architecture Details} \label{sec:arch}

In this section, we present the network architectures used in all of our experiments.

\subsection{Track Proposal Classifier}

Table \ref{table:joint_architecture} shows the network architecture for our joint appearance and motion model with the proposed multi-track pooling module. We used this network to test the proposed approach on the MOT Challenge Benchmarks in Table 6, 7, and 8 of the main paper. Table \ref{table:architecture_2} shows the appearance baseline model used in Table 2, 3, 4, and 5 of the main paper.

\subsection{Track Proposal Classifier Input} For the appearance models, an object image (i.e. object detector output) is used as input. We first resize the object image to $64 \times 128$ (width, height) and then subtract the ImageNet mean from the image. For the motion model, a location (top left corner) and scale (width, height) of a detection bounding box is used as input. The motion model input is thus represented by a 4-dimensional vector $(x^{\text{topleft}}, y^{\text{topleft}}, w, h)$. We normalize the input vector using the image size so the final form of input for the motion model is $(\frac{x^{\text{topleft}}}{\text{image width}}, \frac{y^{\text{topleft}}}{\text{image height}}, \frac{w}{\text{image width}}, \frac{h}{\text{image height}})$.

\subsection{Bounding Box Coordinate Corrector}

Following the bounding box regressor presented in \cite{girshick2014rcnn}, we regress four scalars that corrects the location and scale of the original bounding box from a cropped image. We also have an additional prediction head that classifies false positive detections using the same input image. The network architecture that we used for the box corrector is shown in Table \ref{table:architecture_4}. We utilized the same CNN as the one used in a track proposal classifier. Specifically, once the track classifier was trained, we froze all the CNN weights and then trained the additional linear layers in the two prediction heads from scratch. We used raw DPM detections provided by the MOT16 Challenge organizers as training data to train this module.

\begin{table}[t!]
\centering
\scalebox{0.9}{
\begin{tabular}{|l|} 
\hline
\multicolumn{1}{|c|}{Training Set}  \\ \hline
MOT17 - \{02, 04, 05, 09, 10, 11, 13\} \\
MOT15 - \{PETS09-S2L1, ETH - (Sunnyday, Bahnhof), \\
\quad \quad \quad \quad \ \ TUD - (Campus, Stadtmitte), KITTI-(13, 17)\}, \\
ETH - (Jelmoli, Seq01), KITTI - (16, 19),  \\
PETS09-S2L2, TUD-Crossing, AVG-TownCentre  \\ \hline
\multicolumn{1}{|c|}{Validation Set}  \\ \hline
MOT 19 Challenge - \{01, 02, 03, 05\} \\ \hline
\multicolumn{1}{|c|}{Test Set}  \\ \hline
MOT17 - \{01, 03, 06, 07, 08, 12, 14\} \\ \hline 
\end{tabular}
}
\caption{Split 1} \label{table:splits1}
\end{table}

\begin{table}[t!]
\centering
\scalebox{0.9}{
\begin{tabular}{|l|} 
\hline
\multicolumn{1}{|c|}{Training Set}  \\ \hline
MOT15 - \{PETS09-S2L1, ETH - (Sunnyday, Bahnhof), \\
\quad \quad \quad \quad \ \ TUD - (Campus, Stadtmitte), KITTI-(13, 17)\}, \\
ETH - (Jelmoli, Seq01), KITTI - (16, 19),  \\
PETS09-S2L2, TUD-Crossing, AVG-TownCentre  \\ \hline
\multicolumn{1}{|c|}{Validation Set}  \\ \hline
MOT17 - \{02, 04, 05, 09, 10, 11, 13\} \\ \hline
\multicolumn{1}{|c|}{Test Set}  \\ \hline
MOT17 - \{01, 03, 06, 07, 08, 12, 14\} \\ \hline 
\end{tabular}
}
\caption{Split 2} \label{table:splits2}
\end{table}

\section{Additional Training Details} \label{sec:training_details}

In this section, we describe details about the training settings that we used for our ablation studies.

\begin{table*}[t]
\centering
\scalebox{0.7}{
    \begin{tabular}{|lr||lr||lr||lr||lr|}
    \hline
    \multicolumn{10}{ |c| }{Soft-max \qquad \qquad \quad \quad \ \, \, 2} \\
    \hline
    \multicolumn{10}{ |c| }{FC \qquad \qquad \qquad \qquad \ \ \ \, \, 2 } \\
    \hline
    \multicolumn{10}{ |c| }{FC-relu \qquad \qquad \qquad \ \, \,  \, 24 }\\
    \hline
    \multicolumn{10}{ |c| }{Concatenation \qquad \qquad \, \ 24 }\\
    \hline
    \multicolumn{8}{ |c|| }{ $2 \times$ FC-relu \qquad \qquad \ \ \, \, \ \ 16 } & \multicolumn{2}{ c| }{$2 \times$ FC-relu \, \, \, 8} \\
    \hline
    \multicolumn{8}{ |c|| }{Concatenation ($\mathbf{m}^{\text{all}}_{t}$) \qquad 16 } & \multicolumn{2}{ c| }{} \\
    \hline
    \multicolumn{4}{ |c|| }{Matrix-vector Multiplication-relu ($\mathbf{m}^{+}_t$) \quad \quad \quad \quad \quad \quad 8} & \multicolumn{4}{ c|| }{Max-pooling ($\mathbf{m}^{-}_t$) \quad \quad \quad  \quad  \quad \quad \quad \quad \quad \quad \quad  \quad  \quad \ \quad  \quad \quad \quad \quad \, \ \ \ \ \ $8$} & \multicolumn{2}{ c| }{}\\
    \hline
    \multicolumn{2}{ |c|| }{} & \multicolumn{2}{ c|| }{} & \multicolumn{4}{ c|| }{Matrix-vector Multiplication-relu ($\mathbf{M}^{-}_t$) \quad \quad \quad  \quad  \quad \quad \quad $(M-1) \times 8$}  & \multicolumn{2}{ c| }{}\\
    \hline
    Reshape & $8 \times 256$ &Reshape &  $256 \times 1$ & Reshape & $(M-1) \times 8 \times 256$ &Reshape &  $256 \times 1$ & {} & {} \\ 
    \hline
    LSTM & 2048  & & & LSTM & $(M-1) \times 2048$  & & & FC-relu & 8\\
    \hline
    FC-relu & 256 & FC-relu & 256 & FC-relu & $(M-1) \times 256$ & FC-relu & 256 & LSTM & 64  \\
    \hline
    ResNet-50 & 2048 & ResNet50 & 2048 & ResNet-50 & $(M-1) \times 2048$ & ResNet50 & 2048 & FC-relu & 64 \\
    \hline
    $\mathbf{x}^{+}_{t-1}$ & $128 \times 64 \times 3$ &  $\mathbf{x}_{t}$ & $128 \times 64 \times 3$ & $\mathbf{x}^{-}_{t-1, i}$ & $(M-1) \times 128 \times 64 \times 3$ &  $\mathbf{x}_{t}$ & $128 \times 64 \times 3$ &  $\mathbf{x}^{\text{location, scale}}_{t}$ & 4 \\
    \hline
    \end{tabular}
}
\caption{The appearance + motion model used to generate the results in Table 6, 7, and 8 of the main paper. The number of non-target object tracks used in the multi-track pooling module is represented by $M-1$.} \label{table:joint_architecture}
\end{table*}

\begin{table}[t]
\centering
\scalebox{0.8}{
    \begin{tabular}{|lr||lr|}
    \hline
    \multicolumn{4}{ |c| }{Soft-max \qquad \ \ \, \, 2} \\
    \hline
    \multicolumn{4}{ |c| }{FC \qquad \qquad \, \, \, \ \ 2 } \\
    \hline
    \multicolumn{4}{ |c| }{Matrix-vector Multiplication-relu ($\mathbf{m}_t$) \quad \quad \quad  \quad  \quad \quad \quad 8} \\
    
    \hline
    Reshape & $8 \times 256$ &Reshape &  $256 \times 1$ \\ 
    \hline
    LSTM & 2048 & & \\
    \hline
    FC-relu & 256 & FC-relu & 256  \\
    \hline
    ResNet-50 & 2048 & ResNet50 & 2048 \\
    \hline
    $\mathbf{x}_{t-1}$ & $128 \times 64 \times 3$ &  $\mathbf{x}_{t}$ & $128 \times 64 \times 3$ \\
    \hline
    \end{tabular}
}

\caption{The baseline appearance model that we compared with our proposed appearance model.} \label{table:architecture_2}
\end{table}

\begin{table}[t!]
\centering
\scalebox{0.8}{
    \begin{tabular}{|lr||lr|}
    \hline
    FC & 4 & Soft-max & 2\\
    \hline
    $2 \times$ FC-relu & 512 & $2 \times$ FC-relu & 512\\
    \hline
    \multicolumn{4}{ |c| }{Reshape \qquad \qquad \qquad \qquad \qquad \qquad 16384} \\
    \hline
    \multicolumn{4}{ |c| }{ResNet-50 (Block4) \qquad \, \ \qquad $4 \times 2 \times 2048$} \\
    \hline
    \multicolumn{4}{ |c| }{$\mathbf{x}^{+}_{t-1}$ \qquad \qquad \qquad \qquad \qquad \, $128 \times 64 \times 3$ }\\
    \hline
    \end{tabular}
}
\caption{Bounding Box Coordinate Corrector. This is jointly trained with our track proposal classifier so ResNet-50 is shared with the appearance model.} \label{table:architecture_4}
\end{table}

\subsection{Dataset}

Table \ref{table:splits1} and  \ref{table:splits2} shows the training, validation, and test sets that we used in our experiments. For the final MOT Challenge Benchmark results in Table 6, 7, and 8 of the main paper, the training sequences in Table \ref{table:splits1} were used as the training data to train the model. 

\subsection{Training Setting - Appearance Model} \label{sec:training_setting}

We used the SGD optimizer with the initial learning rate of 0.005 for Bilinear LSTM and the initial learning rate of 0.0005 for ResNet 50 (pre-trained on ImageNet). 
We trained the model with the initial learning rate for the first 4 epochs ($\sim$ 120k iterations), and then reduced the learning rate with the decay factor of 0.1 for the next 4 epochs and reduced it one more time for the last 4 epochs ($\sim$ 360k iterations in total). 

For the actual tracking episodes, we used truncated BPTT with a temporal window size of 10. 
We used all the ground truth tracks in the current frame as our training data so the mini-batch size in this case was equal to the number of ground truth tracks in the current frame. 

For the random tracking episodes, we used 40 frames as the maximum frame gap in the randomly selected start and end frame. Thus, each mini-batch could contain a track whose length is up to 40. Due to the limited GPU memory, the number of tracks for random tracking episodes was limited by  $N_{\text{max}}$. We used $N_{\text{max}}=8$ in our experiments.

\subsection{Training Setting - Motion Model}

We used the Adam optimizer \cite{Kingma15} with the initial learning rate of 0.001. 
We trained the motion model with the initial learning rate for the first 4 epochs ($\sim$ 120k iterations), and then reduced the learning rate with a decay factor of 0.1 for the next 2 epochs and reduced it one more time for the last 6 epochs ($\sim$ 360k iterations in total). 

\subsection{Training Setting - Appearance and Motion Model}

We first trained the appearance and motion models separately as described above before jointly training the model presented in Table \ref{table:joint_architecture}. Thus, new layers (i.e. top five rows in Table \ref{table:joint_architecture}) were trained from scratch, and the rest of the layers (except for ResNet 50 in which all the weights were frozen) was fine-tuned from the pre-trained models. We used the SGD optimizer with the initial learning rate of 0.005 for the new layers and 0.0005 for the pre-trained layers. 
We trained the model with the initial learning rate for the first 2 epochs ($\sim$ 60k iterations), and then reduced the learning rate with the decay factor of 0.1 for the next 2 epochs ($\sim$ 120k iterations in total). 

When we trained the joint model, we realized that the appearance features can be stronger than the motion features in the beginning of the training. As a result, our resulting model heavily relied on the appearance features, often ignoring the motion features. In order to make our model balance between these two types of features, we placed a dropout layer on the appearance features right before the concatenation layer (i.e. top $6^{\text{th}}$ row on the appearance side in Table \ref{table:joint_architecture}). In the beginning of the training, we randomly dropped appearance features and then gradually decreased the drop rate as the training proceeded. This trick prevented the joint model from relying too much on the appearance cues in the early training stage. In our experiments, we used 0.9 as the drop rate for the first $\sim$19k iterations, 0.6 for the next $\sim$10k iterations, 0.3 for the next $\sim$9k iterations, and 0.0 for the rest of the training. 

\subsection{Online Hard Example Mining}

We found that online hard example mining can improve the model performance. We trained all the model with all the training examples for the first two epochs ($\sim$60k iterations). We trained the models for the remaining epochs with top $k$ hard examples (i.e. $k$ examples that incurred high loss values) in the mini-batch. We used $k=30$ in our experiments. Note that the actual tracking episodes that we generated as our training data enabled effective online hard example mining since each mini-batch contained all possible matchings between the tracks and the new detections in the selected frame.

\subsection{Missing Detection Augmentation}

We randomly drop the bounding boxes in the tracks for missing detection augmentation. For each track in each mini-batch, we randomly choose the missing detection rate from a probability between 0.1 and 0.9. After selecting the missing detection rate, we randomly drop the bounding boxes in the selected track according to the selected rate.

\subsection{Noisy Track Augmentation} In addition to using the ground truth tracks as the training data, we also generate tracks from noisy object detections. Given ground truth tracks and noisy detections, one can assign correct track IDs to noisy detections by finding an assignment that maximizes the Intersection over Union score (IoU) between ground truth tracks and object detections. 
The MOT Challenge Benchmark provides public object detections from three detectors (DPM \cite{lsvm-pami}, FRCNN \cite{renNIPS15fasterrcnn}, SDP \cite{Yang_2016_CVPR}). 
Thus, we generate three additional sets of noisy tracks constructed from these public detections. Note that the track-detection assignments are optimal although the resulting tracks are noisier than the original ground truth tracks. Localization and missing detection errors caused by the object detector are embedded naturally in such tracks, which can potentially help the track classifier to generalize better in testing when in noisy detections are used as input to the tracker.

\typeout{get arXiv to do 4 passes: Label(s) may have changed. Rerun}
\end{document}